# Seg-LSTM: Performance of xLSTM for Semantic Segmentation of Remotely Sensed Images

Qinfeng Zhu, *Graduate Student Member, IEEE,* Yuanzhi Cai, *Member, IEEE,* Lei Fan, *Member, IEEE*

*Abstract*— Recent advancements in autoregressive networks with linear complexity have driven significant research progress, demonstrating exceptional performance in large language models. A representative model is the Extended Long Short-Term Memory (xLSTM), which incorporates gating mechanisms and memory structures, performing comparably to Transformer architectures in long-sequence language tasks. Autoregressive networks such as xLSTM can utilize image serialization to extend their application to visual tasks such as classification and segmentation. Although existing studies have demonstrated Vision-LSTM's impressive results in image classification, its performance in image semantic segmentation remains unverified. Our study represents the first attempt to evaluate the effectiveness of Vision-LSTM in the semantic segmentation of remotely sensed images. This evaluation is based on a specifically designed encoder-decoder architecture named Seg-LSTM, and comparisons with state-of-the-art segmentation networks. Our study found that Vision-LSTM's performance in semantic segmentation was limited and generally inferior to Vision-Transformers-based and Vision-Mamba-based models in most comparative tests. Future research directions for enhancing Vision-LSTM are recommended. The source code is available from https://github.com/zhuqinfeng1999/Seg-LSTM.

## I. INTRODUCTION

IMAGE semantic segmentation is a fundamental task in computer vision, involving the partitioning of an image into distinct regions corresponding to specific object classes [1]. This task provides a pixel-level understanding of images, which is crucial for applications requiring precise delineation of object boundaries and detailed scene interpretation, such as medical image analysis, autonomous driving, and remote sensing analysis.

For remotely sensed images acquired from aerial or satellite platforms, their semantic segmentation involves classifying each pixel of an image into predefined categories such as water bodies, buildings, roads, and vegetation [2]. This type of segmentation is essential for applications like urban planning, environmental monitoring, and agricultural management.

In the domain of semantic segmentation of remotely sensed images, deep learning methods have become the mainstream and effective approaches, particularly convolutional neural networks (CNNs) [3] and Vision Transformers (ViTs) [4]. Despite their continuous improvements in accuracy over the years, challenges remain due to the limited receptive field of CNNs and the quadratic complexity of ViTs when handling remote sensing image segmentation.

Similarly, in the domain of large language models, researchers have faced the quadratic complexity limitation of Transformers, which hinders their performance on long-sequence problems. This challenge has led to the development of autoregressive networks, with recent architectures such as xLSTM [5] and Mamba [6, 7] demonstrating exceptional performance in large language models.

These newly developed architectures for large language models can readily be applied to image-related tasks, thanks to the image serialization mechanism proposed with ViT. For example, Vim [8] and VMamba [9] successfully established foundational visual models for Mamba through bidirectional and quad-directional scanning methods, respectively. Vision Mamba has demonstrated impressive performance in numerous tasks in the domains of remote sensing [10], medical imaging and video understanding, highlighting its versatility and effectiveness. Similarly, Vision-LSTM [11] employs an image serialization technique, which alternates between forward and backward scanning to enhance computational efficiency. To date, the efficacy of Vision-LSTM has been validated only on the image classification task using ImageNet. However, according to MambaOut's [12] experiments, image classification tasks do not inherently depend on long-sequence modeling capabilities. Instead, tasks such as segmentation and detection are essential for demonstrating a visual model's ability to handle long sequences. Currently, there is a lack of research addressing this aspect, highlighting the need to validate the effectiveness of Vision-LSTM in semantic segmentation tasks.

This research aims to explore the optimal semantic segmentation architecture and compare it against other high-performing networks. It represents the first application of the Vision-LSTM architecture to semantic segmentation of remotely sensed images. We designed a semantic segmentation framework using an encoder-decoder architecture to explore the effectiveness of the xLSTM architecture. Through extensive experiments, we validated the

This work was supported in part by the Xi'an Jiaotong-Liverpool University Research Enhancement Fund under Grant REF-21-01-003, and in part by the Xi'an Jiaotong-Liverpool University Postgraduate Research Scholarship under Grant FOS2210JJ03. *(Corresponding author: Lei Fan.)*

Qinfeng Zhu and Lei Fan are with the Department of Civil Engineering, Xi'an Jiaotong-Liverpool University, Suzhou, 215123, China. (e-mail: Qinfeng.Zhu21@student.xjtlu.edu.cn; lei.fan@xjtlu.edu.cn)

Yuanzhi Cai is with the CSIRO Mineral Resources, Kensington, WA 6151, Australia. (e-mail: Yuanzhi.Cai@CSIRO.AU)

impact of different decoders and multi-level network depths, and subsequently identified an optimal xLSTM-based framework for segmentation of remotely sensed images. Furthermore, we compared it with representative methods.

The main contributions of this paper are as follows: (i) This paper is the first to apply xLSTM to image semantic segmentation tasks, validating its effectiveness on high-resolution remote sensing datasets. (ii) Through extensive experiments, we explored the optimal architecture for the xLSTM semantic segmentation framework. (iii) By conducting a comprehensive comparison with CNN-based, ViT-based, and Mamba-based methods, this paper provides a discussion on xLSTM-based semantic segmentation methods and outlines directions for future work.

## II. THE SEG-LSTM FRAMEWORK

This section provides an in-depth explanation of our designed implementation of Vision-LSTM to segmentation, named Seg-LSTM. We begin by introducing the xLSTM and Vision-LSTM architectures, followed by a comprehensive overview of the Seg-LSTM model, structured as an encoder-decoder framework.

### A. Preliminaries

LSTM [13], proposed by Hochreiter and Schmidhuber in 1997, is an improved architecture for RNNs that alleviates the problem of capturing long-term dependencies. This method introduces gating units, including input, forget, and output gates, to control the flow of information. These gates allow LSTM to remember or forget information, thus better handling long-term dependencies. LSTM networks have been foundational in many sequential tasks and have been integrated into hybrid models with CNNs in recent methodological advancements. However, LSTM has some inherent issues. Firstly, due to its sequential nature during training, it is difficult to parallelize, resulting in low efficiency. Secondly, LSTM still struggles with long-distance dependencies when dealing with extremely long sequences. Consequently, LSTM is less suitable for complex tasks like large language models, where Transformers have become the preferred choice.

Recently developed xLSTM [5] makes improvements in the gating mechanism and memory structure. This approach introduces an exponential gating mechanism, offering dynamic information filtering capabilities. It also introduces sLSTM and mLSTM memory cells. sLSTM introduces a scalar update mechanism to improve robustness in long sequences, while mLSTM extends vector operations to matrix operations, addressing the parallelization issue and enhancing computational efficiency. These units, connected via residual connections, constitute the xLSTM architecture. Extensive experiments have demonstrated that xLSTM performs comparably to Transformers in large language models.

xLSTM can readily be applied to images with image serialization such as that proposed for ViT and scanning strategies such as bi-directional scanning in Vim [8] and quad-directional scanning in VMamba [9], leading to the realization of Vision-LSTM [11]. Unlike Vim's and VMamba's scanning strategies, Vision-LSTM proposes a more efficient alternating bi-directional scanning method. Through extensive experiments on the ImageNet dataset, Vision-LSTM's Tiny and Small models have shown more competitive results compared to Vim, demonstrating its potential as a visual backbone network. Therefore, it is meaningful to test its performance in downstream segmentation tasks as a visual backbone network.

### B. Seg-LSTM

Seg-LSTM is designed using Vision-LSTM as the backbone within an encoder-decoder framework, as illustrated in Figure 1. Unlike Vision-LSTM, which serially connects several blocks, Seg-LSTM extracts features in four stages, performing semantic segmentation by integrating multi-level features with the decoder, as illustrated in Figure 1. Specifically, an input image is first processed by the Stem module, which linearly projects the image into non-overlapping patches. Learnable position embeddings are then added to each patch token, consistent with the approach used in ViT. Next, these tokens are passed through several ViL blocks across the four stages. The features from these stages are then selectively fed into the decoder, based on different decoder designs, to complete the segmentation task.

The design of the ViL Block is similar to the Mamba Block [7], both utilizing a Gated MLP architecture, as depicted on the right side of Figure 1. However, the scanning method is chosen based on the parity of the block. Specifically, the ViL Block is a residual network with skip connections, featuring two main branches. One branch determines the parity of the current block: odd-numbered blocks perform forward scanning and compute through the mLSTM layer, while even-numbered blocks perform backward scanning and computation. The other branch includes a linear mapping, followed by a SiLU non-linear activation.

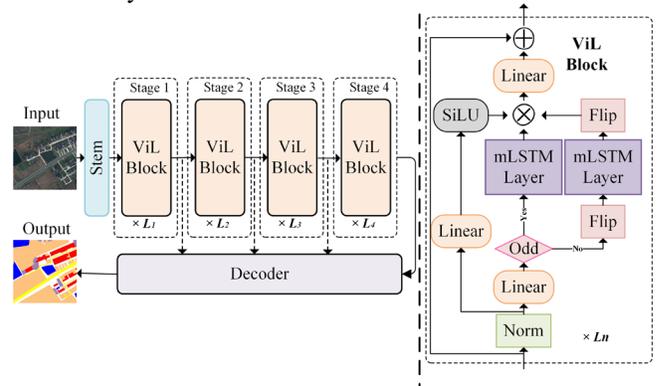

**Fig. 1.** Overall architecture of Seg-LSTM and the detailed architecture of the ViL block.

III. Experimental Setup

*A. Datasets and Metrics*

To comprehensively test the effectiveness of xLSTM in semantic segmentation of remotely sensed images, we conducted numerous experiments using three benchmark datasets: LoveDA [14], ISPRS Vaihingen, and ISPRS Potsdam.

The LoveDA dataset comprises 1669 validation images, 1796 test images, and 2522 training images, each with a resolution of 1024×1024 pixels and a spatial resolution of 30 cm. These images span seven categories: background, buildings, roads, water, barren areas, forests, and agricultural lands. The validation set is employed to assess performance in this study.

The ISPRS Vaihingen dataset is composed of 33 high-resolution aerial images, each with a spatial resolution of 9 cm and varying sizes, averaging around 2494×2064 pixels. These images encompass near-infrared, red, and green spectral bands and are classified into six categories: impervious surfaces, buildings, low vegetation, trees, cars, and clutter. For training purposes, images labeled with IDs 1, 3, 5, 7, 11, 13, 15, 17, 21, 23, 26, 28, 30, 32, 34, and 37 are utilized, whereas the remaining 17 images serve as the validation set.

The ISPRS Potsdam dataset includes the same categories as ISPRS Vaihingen but provides a finer spatial resolution of 5 cm. This dataset comprises 38 images, each uniformly sized at 6000×6000 pixels, and the RGB channels are used in this study. Training images are identified by IDs 2_10, 2_11, 2_12, 3_10, 3_11, 3_12, 4_10, 4_11, 4_12, 5_10, 5_11, 5_12, 6_07, 6_08, 6_09, 6_10, 6_11, 6_12, 7_07, 7_08, 7_09, 7_10, 7_11, and 7_12, with the remaining 14 images designated for validation. Similar to ISPRS Vaihingen, the clutter category is excluded from evaluation.

In this study, we utilize the mean Intersection over Union ($mIoU$) metric to evaluate the performance of semantic segmentation models. mIoU is a widely adopted evaluation criterion in semantic segmentation of remotely sensed images, measuring the accuracy of predicted segmentation against ground truth by calculating the average IoU across all classes. The IoU for a particular class is defined as the ratio of the intersection of the predicted and ground truth areas to their union, formulated in Eqn. (1).

$$IoU = \frac{\text{Area of Overlap}}{\text{Area of Union}} \qquad (1)$$

Mathematically, for $N$ classes, mIoU is expressed in Eqn (2).

$$mIoU = \frac{1}{N} \sum_{i=1}^{N} \frac{TP_i}{TP_i + FP_i + TN_i} \qquad (2)$$

where $TP_i$, $FP_i$, and $TN_i$ represent the true positive, false positive, and false negative counts for class $i$, respectively.

mIoU provides an assessment of the segmentation accuracy by considering both over-segmentation and under-segmentation errors, thereby serving as a robust indicator of model performance.

*B. Training Settings*

TABLE I
TRAINING SETTING FOR SEMANTIC SEGMENTATION NETWORKS ON THREE DATASETS

| Dataset | LoveDA | Vaihingen | Potsdam |
|---|---|---|---|
| Resize | 2048×512 | 512×512 | 1024×1024 |
| Patch size | 512×512 | | |
| Total training | 15000 iterations | | |
| Batch size | 16 | | |
| Optimizer | AdamW | | |
| Weight decay | 0.01 | | |
| Schedule | PolyLR | | |
| Warmup | 1500 iterations | | |
| Learning rate | 0.0006 | | |
| Loss function | Cross entropy loss | | |

Determined through extensive experimental tuning, optimal experimental settings used in our tests are summarized in Table I. To alleviate the overfitting problem during training, we used a range of data augmentation methods [15], including photometric distortion, random resize, random crop, and random flip. For the backbone network Vision-LSTM, we utilized a total of 24 blocks with a dimension of 384. To ensure fairness, all methods were trained using a fully supervised approach. All experiments were conducted on two RTX 3090 and two RTX 4090D GPUs.

*C. Experiments Design*

In the experimental design, we experimented with different decoders and multi-stage network depth to optimize the performance of Seg-LSTM. Our study considers several representative and high-performing decoders, including UperNet, DeepLabV3, DeepLabV3+, APCNet, and ANN, to understand their effectiveness when combining Vision-LSTM as the backbone network. UperNet employs a Feature Pyramid Network (FPN) to effectively fuse multi-scale features, enhancing spatial information and contextual understanding. DeepLabV3 incorporates an Atrous Spatial Pyramid Pooling (ASPP) module, which excels at capturing multi-scale context through multiple atrous convolutions, while DeepLabV3+ provides various optimizations to ASPP. APCNet uses an adaptive context aggregation mechanism to dynamically adjust the receptive field, effectively aggregating contextual information. ANN introduces attention mechanisms to capture long-range dependencies, improving the network's ability to model complex patterns and relationships in the data.

TABLE II
EFFECTS OF DEPTH OF EACH STAGE AND DECODERS ON SEGMENTATION ACCURACY. THE ACCURACY IS PRESENTED BY THE
mIoU METRIC. THE HIGHEST SCORES ARE HIGHLIGHTED IN BOLD.

| Depth | Encoder | Decoder | Potsdam | Vaihingen | LoveDA |
|---|---|---|---|---|---|
| 6-6-6-6 | Vision-LSTM | UperNet | **75.24** | 62.66 | **37.80** |
| | | DeepLabV3 | 73.31 | 62.33 | 35.59 |
| | | DeepLabV3+ | 67.73 | **62.88** | 35.59 |
| | | APCNet | 73.62 | 61.57 | 36.80 |
| | | ANN | 72.97 | 62.94 | 35.04 |
| 4-4-12-4 | Vision-LSTM | UperNet | **75.43** | **63.25** | 37.47 |
| | | DeepLabV3 | 72.96 | 62.26 | **38.72** |
| | | DeepLabV3+ | 64.46 | 62.43 | 34.75 |
| | | APCNet | 73.75 | 62.20 | 38.07 |
| | | ANN | 72.95 | 63.06 | 38.04 |

TABLE III
ACCURACY OF SEMANTIC SEGMENTATION RESULTS FROM SEG-LSTM AND OTHER COMPARED METHODS. THE ACCURACY
IS PRESENTED BY THE mIoU METRIC. THE HIGHEST SCORES ARE HIGHLIGHTED IN BOLD.

| | Encoder | Decoder | Potsdam | Vaihingen | LoveDA | #Parms (M) |
|---|---|---|---|---|---|---|
| CNN-based | ConvNext | UperNet | 74.70 | 67.42 | 36.81 | 59.20 |
| | ResNet50 | UperNet | 74.98 | 70.25 | 32.86 | 64.00 |
| | ResNet50 | DeepLabV3+ | 75.23 | 69.10 | 34.60 | 41.20 |
| ViT-based | Swin-T | UperNet | 76.46 | 71.72 | 41.08 | 58.90 |
| | Mix ViT | Segformer | 81.13 | 70.23 | 43.16 | - |
| Mamba-based | Samba | UperNet | **82.29** | 73.56 | 47.11 | 51.90 |
| | VMamba | UperNet | 81.73 | **76.22** | **47.82** | 64.29 |
| Seg-LSTM | Vision-LSTM | UperNet | 75.24 | 62.66 | 37.80 | 51.80 |

Additionally, this study examines the impact of block distribution across the four stages of the encoder on performance. In semantic segmentation encoders, it is common to divide the network into four stages. Initially, we use an equal number of blocks in each stage, with each stage containing 6 blocks. Inspired by the designs of Swin Transformer and VMamba, we then adjust the network depth in the third stage to study its impact, specifically configuring the four stages to contain 4, 4, 12, and 4 blocks, respectively. This approach aims to enhance the model's ability to capture more abstract and complex patterns and structures in the image, thereby deepening the understanding on images.

## IV. RESULTS

### A. Impact of Decoders and Multistage Depth

The performance of the Seg-LSTM architecture with different decoders and varying network depths across the four stages on the LoveDA, ISPRS Vaihingen, and ISPRS Potsdam datasets is presented in Table II.

Upon examination, it can be observed that the variation in network depth across different stages did not significantly impact the segmentation results. Both strategies, whether using an equal distribution of blocks (6-6-6-6) or an adjusted depth (4-4-12-4), were feasible. Therefore, in subsequent experiments, we used the configuration of 6 blocks per stage.

However, the choice of decoders exhibited a more substantial effect. Overall, UperNet proved to be the most suitable decoder compared to other considered decoders, exhibiting stable and outstanding performance across all three datasets. DeepLabV3 and DeepLabV3+ showed superior performance on specific datasets, but this superiority in performance was not consistently over all datasets and network depths.

### B. Comparison with SOTA methods

Table III shows the comparison of the performance of the Seg-LSTM architecture with those achieved by CNN-based, ViT-based, and Mamba-based architectures, on the LoveDA, ISPRS Vaihingen, and ISPRS Potsdam datasets, along with the corresponding parameter counts. The tabulated results show that Seg-LSTM outperformed CNN-based methods on the ISPRS Potsdam and LoveDA datasets, however its overall performance still fell short compared to ViT-based and Mamba-based methods.

As demonstrated in Table III, Mamba-based methods showed excellent performance in semantic segmentation even with unidirectional scanning [16]. However, Vision-LSTM, which adopts an alternating unidirectional strategy to control computational costs, did not achieve comparable performance. This simple scanning strategy might have affected Vision-LSTM's global image modeling capabilities. Therefore, integrating multi-directional scanning, similar to those adopted in Vim [8] and VMamba [9] could potentially enhance Vision-LSTM's performance.

## V. FUTURE WORK

While Vision-LSTM has demonstrated good performance in image classification tasks, this does not validate its long-sequence modeling capabilities in all visual tasks. Our designed Seg-LSTM architecture, through extensive experiments, revealed that Vision-LSTM's performance was inferior to ViT-based and Mamba-based methods for semantic segmentation of remotely sensed images.

Based our study and understanding, the following future research directions are suggested: (i) Both VMamba and Swin Transformer employ downsampling in the four feature extraction stages of their encoders. This staged downsampling enables better extraction of multi-scale features while reducing computational complexity. This approach is worth considering for future integration into the Vision-LSTM encoder architecture. (ii) Vision-LSTM uses an alternating unidirectional scanning strategy, which reduces computational complexity but sacrifices global understanding of images. Exploring multi-directional scanning strategies in vision-LSTM for image semantic segmentation, while maintaining computational efficiency, is a valuable area for further research. (iii) This study did not consider the transferability of pretrained encoders to downstream segmentation tasks. Given Vision-LSTM's strong performance on ImageNet image classification, exploring fine-tuning methods for pretrained models is a meaningful direction for future research. (iv) Significant progress has been made with Mamba in U-Net architectures. Further exploration of using Vision-LSTM within U-Net architectures, especially for small-sample semantic segmentation, is recommended.

## VI. Conclusion

In this study, we explored the effectiveness of Vision-LSTM for semantic segmentation of remote sensing imagery. To this end, we designed an encoder-decoder framework named Seg-LSTM, and experimented with different decoders and varying encoder network depths to optimize its performance. We found that UperNet was the most effective decoder, while the considered network depths exhibited minimum impact on the performance of Seg-LSTM.

Quantitative comparisons with CNN-based, ViT-based, and Mamba-based methods revealed that Vision-LSTM's performance in long-sequence task modeling was limited and generally inferior to most ViT-based and Mamba-based methods, indicating significant room for improvement. This is likely due to the alternating scanning method of Vision-LSTM for image sequences, which influences its global sequence modeling. Our study suggests that the architectural design, computational efficiency, and transferability of Vision-LSTM merit further in-depth research.
5